\documentclass{cas-sc}
\usepackage[authoryear]{natbib}
\usepackage{graphicx}
\usepackage{subcaption}
\usepackage{xcolor}
\usepackage{setspace}
\usepackage{lineno}

\ExplSyntaxOn \cs_gset:Npn \__first_footerline: { \group_begin: \small \sffamily \__short_authors: \group_end: } \ExplSyntaxOff

\begin{document}
\let\printorcid\relax

\shorttitle{Tabular foundation models in soil spectroscopy}
\shortauthors{Barkov et al.}

\title[mode = title]{From field-scale to large-scale spectral libraries: Tabular foundation models in soil spectroscopy}

\author[1,2]{Viacheslav Barkov}
\author[1,2]{Jonas Schmidinger}
\author[2]{Robin Gebbers}
\author[1,3]{Martin Atzmueller}

\affiliation[1]{organization={Osnabrück University, Joint Lab Artificial Intelligence and Data Science}, city={Osnabrück}, country={Germany}}
\affiliation[2]{organization={Leibniz Institute for Agricultural Engineering and Bioeconomy (ATB), Department of Agromechatronics}, city={Potsdam}, country={Germany}}
\affiliation[3]{organization={German Research Center for Artificial Intelligence (DFKI), Cooperative and Autonomous Systems (CAS)}, city={Osnabrück}, country={Germany}}

\begin{abstract}
  Visible and near-infrared (vis-NIR) and mid-infrared (MIR) spectroscopy enable rapid, cost-effective prediction of soil properties. Yet, translating high-dimensional, highly collinear spectra into accurate soil property predictions remains challenging, particularly when employing machine learning. We systematically investigated regression models and dimensionality reduction approaches for spectroscopic modeling across 85 soil property prediction tasks from open benchmark datasets in pedometrics, spanning field-scale digital soil mapping and a global soil spectral library. We compared an in-context learning tabular foundation model (TabPFN), a convolutional neural network (CNN), rule-based regression (Cubist), Random Forest, and partial least squares regression (PLSR) using full spectra as well as features derived from principal component analysis (PCA) and partial least squares (PLS) latent variables. TabPFN consistently delivered the best overall performance across spectral regions, properties, and scales, including large spectral library tasks with tens of thousands of soil samples. Notably, TabPFN applied directly to full spectra already surpassed all classical baselines, showing that explicit dimensionality reduction is not strictly required for strong performance. Further improvements were achieved through PLS, which proved to be an effective dimensionality reduction strategy for all models. Combining PLS latent variables with TabPFN yielded the best predictions overall, lowering the root mean squared error (RMSE) relative to the strongest classical baseline on 44 of the 46 large-scale spectral library tasks, by a median of 14\%. Our findings provide evidence-based guidance for spectroscopic calibration model selection across operational scales, demonstrating that the long-standing advantages of PLSR and modern tabular foundation models complement each other in chemometrics.
\end{abstract}


\nonumnote{Notice: The published version of this article is available in \textit{Geoderma} at: \href{https://doi.org/10.1016/j.geoderma.2026.118138}{https://doi.org/10.1016/j.geoderma.2026.118138}}

\maketitle

\onehalfspacing

\section{Introduction}

Diffuse reflectance spectroscopy enables the rapid and non-destructive estimation of material properties from their characteristic molecular absorption signatures~\citep{frei2019diffuse}. It is therefore widely used as a cost-effective analytical technique across various environmental disciplines, including soil science. In the infrared region in particular, certain soil minerals and organic compounds exhibit distinct absorption features arising from the vibrations of their chemical bonds~\citep{stoner1981characteristic,ben1999soil}. How well soil properties can be estimated from these features depends on the spectral region. In the mid-infrared (MIR) region (approximately 4{,}000--400~cm$^{-1}$), the fundamental molecular vibrations of soil constituents produce numerous, intense, and specific absorption bands. In the visible and near-infrared (vis-NIR) region (approximately 400--2{,}500~nm), by contrast, absorption is dominated by electronic transitions at visible wavelengths and by weaker overtones and combinations of these fundamental vibrations~\citep{viscarrarossel2006visible}. In both regions, however, the resulting spectral fingerprint contains information about the soil's composition, from which soil properties such as texture, organic carbon, pH, and concentrations of certain nutrients can be inferred~\citep{viscarrarossel2006visible}. Accordingly, both vis-NIR and MIR spectroscopy are increasingly used ex situ in soil laboratories as a substitute for costly laboratory analyses~\citep{nocita2015soilspectroscopy} as well as in situ in the field through proximal soil sensing (PSS) for high-resolution soil mapping~\citep{soriano-disla2014performance,gebbers2019proximal}.

Translating spectral signals into soil property predictions requires robust predictive models, which act as the critical backbone of soil spectroscopy~\citep{nocita2015soilspectroscopy} and combine insights from chemometrics, the application of mathematical and statistical methods to chemical data~\citep{brown2020chemometrics}, and pedometrics, the application of mathematical and statistical methods in soil science~\citep{webster1994pedometrics}. Knowledge from these disciplines enables the efficient estimation of soil properties, a prerequisite for many applications in soil science, such as digital soil mapping (DSM)~\citep{mcbratney2003dsm,minasny2016dsm}.

Owing to its flexibility in modeling the relationship between dependent and independent variables, machine learning (ML) has become the predominant approach for developing such predictive models in soil spectroscopy~\citep{rossel2022diffuse} and for pedometric modeling in general \citep{wadoux2020machine}. In soil spectroscopy specifically, these models operate on a structured, tabular representation of reflectance spectra, placing the task on common methodological ground with numerous other scientific domains where tabular ML is prevalent. However, soils present unique challenges for ML modeling. Soils exhibit great physical and structural complexity~\citep{Jenny1941factors}, with highly heterogeneous composition varying across spatial and temporal scales: soil properties fluctuate at the microscopic scale due to complex pore structures and mineralogical diversity, at the macro and field scales due to environmental and anthropogenic factors, and over time due to high temporal dynamics in properties such as moisture and nutrient content~\citep{herrick2023practical,young2008microbial}.

Moreover, the specifics of diffuse reflectance spectra pose modeling challenges of their own. Spectral measurements can be strongly affected by particle-size heterogeneity and light scattering, by the overlap of absorption features arising from multiple interacting soil constituents, and by variability in water content and other environmental factors when acquired in situ~\citep{ben1999soil,soriano-disla2014performance}. As a consequence, spectra almost never respond to a single soil property unambiguously~\citep{gebbers2019proximal}. In addition, spectrometers output hundreds to thousands of spectral variables~\citep{wang2022spectral}, creating high-dimensional feature spaces. At the field scale, where financial and practical constraints typically limit the number of soil samples~\citep{soderstrom2016adaptation,schmidinger2024effect}, this yields unfavorable feature-to-sample ratios and the curse of dimensionality~\citep{dossantos2023improving,correa2025overfitting}. At larger scales, where national and global soil spectral libraries may contain tens of thousands of samples, the high dimensionality instead introduces substantial computational overhead~\citep{viscarrarossel2016global,gamagedara2025application}. On top of that, neighboring spectral bands are highly correlated, further complicating the application of ML in soil spectroscopy~\citep{dossantos2023improving,wang2022determination,wang2022spectral}.

These challenges have historically shaped a distinct modeling landscape for soil spectroscopy, favoring approaches that differ from standard tabular ML practice. Despite its origins in the 1970s~\citep{wold1975soft}, partial least squares (PLS) regression (PLSR) remains one of the most widely used methods for quantitative spectral analysis of soil to this day \citep[e.g.,][]{gyawali2025measuring,reyes2024spectral,pace2024soil}. Its longevity is well founded: PLSR is exceptionally well suited to spectral data, as it jointly addresses dimensionality reduction and regression while inherently handling multicollinearity and noise~\citep{soriano-disla2014performance,viscarrarossel2006visible,janik1998can}. Convolutional neural networks (CNNs) have more recently attracted interest for soil spectroscopy \citep[e.g.,][]{huang2025using,ng2019convolutional}, as the ordered nature of spectral wavelengths and their high intercorrelation are well suited to convolutional kernels. While these approaches handle dimensionality within the model itself, an alternative strategy is to address it explicitly through dimensionality reduction~\citep[e.g.,][]{correa2025overfitting,safanelli2025open,schmidinger2025limesoda} or feature selection as a preprocessing step~\citep[e.g.,][]{rodriguez-albarracin2024soil,cui2026improving}. This offers flexibility in the choice of regression model and enables practitioners to leverage advances in general-purpose tabular ML as they arise.

Among dimensionality reduction techniques in soil spectroscopy, principal component analysis (PCA) is a well-established choice~\citep{correa2025overfitting,safanelli2025open,schmidinger2025limesoda}. Its appeal is clear: PCA is simple, well-understood, and can be paired with any downstream regression model. Yet the persistent popularity of PLSR in soil spectroscopy, despite decades of ML advances, suggests that there are distinct advantages in the way PLS compresses spectral features. Similar to PCA, PLS decomposes spectra into a reduced set of latent variables. Unlike PCA, however, PLS is a supervised method: the decomposition maximizes the covariance between spectra and the soil property of interest, thereby producing latent variables that are inherently aligned with the prediction target. Extending such target-aware dimensionality reduction beyond the PLSR framework presents a promising opportunity: rather than restricting the extracted PLS latent variables to a linear model, as is done in PLSR, they could be used as input features for modern regression models, effectively combining target-aware dimensionality reduction with the capacity of state-of-the-art algorithms. Despite the intuitive appeal of this idea, it has never been systematically evaluated in soil spectroscopy, having appeared as a dimensionality reduction choice only in a few isolated case studies~\citep[e.g.,][]{zhao2024integrating,liu2017combining,mouazen2010comparison}. Furthermore, whether it could enhance the performance of recent advances in pedometric modeling within soil spectroscopy remains an open question.

A recent development in tabular ML that holds particular relevance for pedometric modeling is the emergence of in-context learning foundation models for tabular data~\citep{hollmann2025accurate,hollmann2023tabpfn,muller2022transformers}. Unlike conventional ML approaches that require dataset-specific training and hyperparameter optimization, these models are pre-trained on large collections of synthetic tabular tasks, and perform inference by conditioning on a labeled training set provided as a context at prediction time~\citep{hollmann2023tabpfn}. TabPFN (Tabular Prior-data Fitted Network) is a pioneer of this paradigm~\citep{hollmann2025accurate,hollmann2023tabpfn}, and has demonstrated strong performance on tabular datasets across diverse domains~\citep{erickson2025tabarena,grinsztajn2025tabpfn}, including state-of-the-art results in field-scale DSM~\citep{barkov2026modern}. Earlier versions of TabPFN, however, were constrained to datasets with limited numbers of samples and features~\citep{hollmann2025accurate,ye2025closer}, creating a direct conflict with the high dimensionality characteristic of soil spectroscopy. To address this limitation, combining PCA with TabPFN was proposed as a workaround both in DSM~\citep{barkov2026modern} and in general tabular ML~\citep{ye2025closer}. Recent versions of tabular foundation models address these limits. TabPFN-3~\citep{grinsztajn2026tabpfn} has demonstrated state-of-the-art performance on general tabular benchmarks with datasets of up to 100{,}000 training samples and nearly 2{,}000 features~\citep{erickson2025tabarena}, aligning well with the needs of soil spectroscopy beyond the field scale and towards the scale of national and global soil spectral libraries. Initial exploratory studies in pedometrics suggest that these advances extend the potential of TabPFN beyond the field scale: \citet{huang2026zeroshot} reported strong accuracy and generalization on subsets of a national soil spectral library of up to roughly 10{,}000 soil samples. However, whether these strengths hold under the specific challenges of larger-scale spectroscopic modeling remains an open question, both when applied directly to spectral data and when combined with typical dimensionality reduction strategies, warranting systematic investigation.

While investigating these questions is essential, it is equally important to do so in a generalizable and fair manner. Current benchmarking practice in pedometrics overwhelmingly relies on single-dataset evaluations~\citep{schmidinger2025limesoda}, even though single-dataset conclusions are vulnerable to dataset-specific biases and often fail to generalize~\citep{boulesteix2017towards,niessl2022over,schmidinger2025limesoda}. This is particularly consequential in soil spectroscopy, where soil surveys are conducted with different objectives and produce datasets with fundamentally different characteristics \citep{chabrillat2019imaging}. At field-scale, where individual fields or farms of roughly one to a few hundred hectares are surveyed, a small number of highly localized soil samples, typically tens to few hundreds \citep{schmidinger2025limesoda}, are obtained to create high-resolution maps for the needs of precision agriculture. In national and global soil spectral libraries, which are compiled for environmental monitoring, large numbers of soil samples are collected with sparse spatial distribution, resulting in large datasets with high diversity and heterogeneity of soil properties. A method that excels in one of these scenarios may not retain its advantages in the other, making cross-scale evaluation necessary to draw transferable conclusions. Further, to ensure fairness and reproducibility, it is necessary to employ openly accessible benchmark dataset collections~\citep[e.g.,][]{schmidinger2025limesoda,safanelli2025open}.

In this study, we systematically investigate modeling strategies and dimensionality reduction approaches for soil spectroscopy across multiple datasets and operational scales. We evaluate whether the demonstrated strengths of TabPFN-3 extend to the specific challenges of spectroscopic modeling, including the first application to large-scale soil spectral library tasks involving tens of thousands of soil samples. We further investigate whether the predictive performance of pedometric models can be enhanced by employing PLS latent variables as input features for both classical algorithms and tabular foundation models. We compare our findings against established baselines and across datasets that span field-scale to global spectral libraries, with openly accessible data and code to ensure fairness and generalizability of conclusions. We additionally investigate whether our findings hold across spectral regions and soil properties with different spectral mechanisms, from spectrally active constituents to properties that are only indirectly related to the spectra. As a result of our study, we aim to provide evidence-based guidance for practitioners selecting modeling strategies for soil spectroscopy across operational scales.

\section{Materials and methods}

\subsection{Datasets}

\subsubsection{Field-scale spectroscopy data}

\begin{figure}[pos=htbp]
  \centering
  \includegraphics[width=0.5\textwidth]{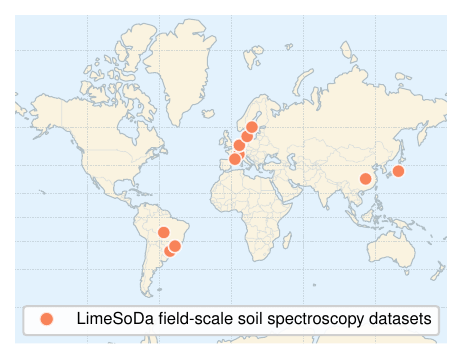}
  \caption{Spatial distribution of the 13 field-scale soil spectroscopy datasets from the LimeSoDa benchmark collection employed in this study. The datasets span seven countries across three continents, encompassing diverse geological settings, spectrometer instruments, study area sizes, and measurement protocols. Base map source: Natural Earth.}
  \label{figure:datasets-map-limesoda}
\end{figure}

The high dimensionality of spectral features combined with the small sample sizes typical of field-scale soil surveys commonly leads to unfavorable feature-to-sample ratios and complicates modeling~\citep{dossantos2023improving,correa2025overfitting}. To draw generalizable conclusions about field-scale soil spectroscopy modeling strategies, we employed the Precision Liming Soil Datasets (LimeSoDa)~\citep{schmidinger2025limesoda}, an open-access benchmark dataset collection of field- and farm-scale high-resolution DSM datasets that aggregates data from multiple independent soil surveys across diverse geographical and geological settings.

For the purpose of our investigation, we employed all LimeSoDa datasets that include features derived from soil spectroscopy, specifically from vis-NIR and MIR spectroscopy. For datasets that additionally contained features derived from other proximal soil sensors, such as apparent electrical resistivity, we retained only the spectroscopic features. This selection resulted in 13 field-scale datasets: eight produced with vis-NIR spectroscopy and five with MIR spectroscopy. Each resulting dataset includes three target soil properties: soil organic carbon (SOC) or soil organic matter (SOM), pH, and clay content. These ground truth target soil properties were determined through wet chemistry laboratory analyses. Since each dataset contains three target soil properties, this resulted in 39 separate soil property prediction tasks. Further details can be found in Supplementary Table~S1 and in \citet{schmidinger2025limesoda}.

\begin{figure}[pos=htbp]
  \centering
  \centering\includegraphics[width=0.5\textwidth]{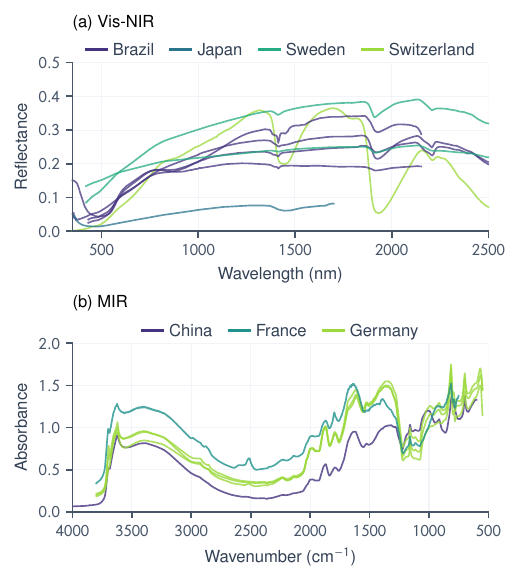}
  \caption{Mean spectra of the individual spectroscopy datasets from LimeSoDa: (a) eight vis-NIR datasets, and (b) five MIR datasets. Each line is the per-band mean spectrum across all soil samples of one dataset. Colors show the origin country of the individual datasets. Vis-NIR spectra are shown as reflectance against wavelength, and MIR spectra as absorbance against wavenumber.}
  \label{figure:datasets-spectra-limesoda}
\end{figure}

The 13 selected datasets encompass a wide range of geological settings, spectrometer instruments, study area sizes, and measurement protocols. The datasets span seven countries across three continents, with study sites in Brazil, China, France, Germany, Japan, Sweden, and Switzerland (see Figure~\ref{figure:datasets-map-limesoda}). Geological settings range from glacial and fluvioglacial deposits, Cretaceous marls with aeolian and fluvial overburden, and Pleistocene periglacial slope deposits in Europe, to heavily weathered tropical soils developed on sandstone, volcanic rock, and diabase. Spectroscopic measurements were obtained using a variety of instruments, including laboratory spectrometers (e.g., Bruker Tensor 27 HTS-XT, Bruker Optik, Ettlingen, Germany) and portable handheld devices deployed in the field (e.g., Agilent 4300, Agilent Technologies, Santa Clara, USA). Study area sizes range from 0.6 to 473~ha. Dataset sizes are typical for field-scale soil surveys, ranging from 32 to 460~soil samples. The vis-NIR spectra span the approximate spectral range of 350--2{,}500~nm, resulting in datasets with up to 2{,}151 spectral variables, while MIR spectra span approximately 550--5{,}400~cm$^{-1}$, resulting in datasets with up to 2{,}489 spectral variables. The mean spectral signatures of individual vis-NIR and MIR datasets are presented in Figure~\ref{figure:datasets-spectra-limesoda}. All datasets are openly accessible under a permissive license (see \hyperref[section:code-data-availability]{Code and data availability}).

\subsubsection{Large-scale soil spectral library}

\begin{figure}[pos=htbp]
  \centering
  \includegraphics[width=0.5\textwidth]{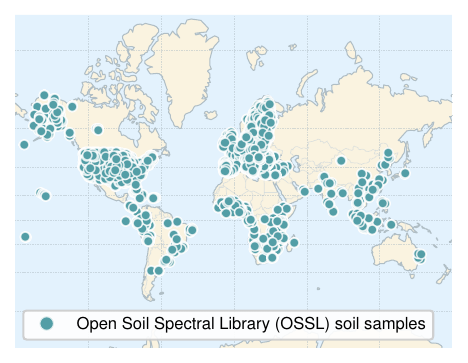}
  \caption{Spatial distribution of soil samples in the Open Soil Spectral Library (OSSL) employed in this study. The OSSL aggregates harmonized spectral data from multiple national and international soil spectral collections, providing near-global spatial coverage. Base map source: Natural Earth.}
  \label{figure:datasets-map-ossl}
\end{figure}

National and global soil spectral libraries compile large numbers of samples spanning diverse soil types, climatic zones, and parent materials~\citep[e.g.,][]{safanelli2025open,fohrafellner2026austrian,meszaros2025vis}. Their larger spatial scale results in greater heterogeneity in soil composition, and the aggregation of data across laboratories with different methods and instruments introduces additional variability~\citep{safanelli2025open}. The high dimensionality of spectral data additionally introduces substantial computational overhead at this scale~\citep{viscarrarossel2016global,gamagedara2025application}.

We employed the Open Soil Spectral Library (OSSL) for the large-scale spectral library evaluation~\citep{safanelli2025open}. OSSL is an open-access global soil spectral library compiled by the Soil Spectroscopy for Global Good initiative that aggregates and harmonizes soil spectral data from multiple independent libraries with open or compatible data-sharing policies. Ground truth target soil properties were determined through wet chemistry laboratory analyses. Spectral data were harmonized to common ranges and resolutions, and laboratory reference data were harmonized by mapping analytical methods to standardized guides and converting shared properties to unified target variables via published transformation rules~\citep{safanelli2025open}. The resulting database contains over 135{,}000 entries with near-global spatial coverage (see Figure~\ref{figure:datasets-map-ossl}).

\begin{figure}[pos=htbp]
  \centering
  \centering\includegraphics[width=0.5\textwidth]{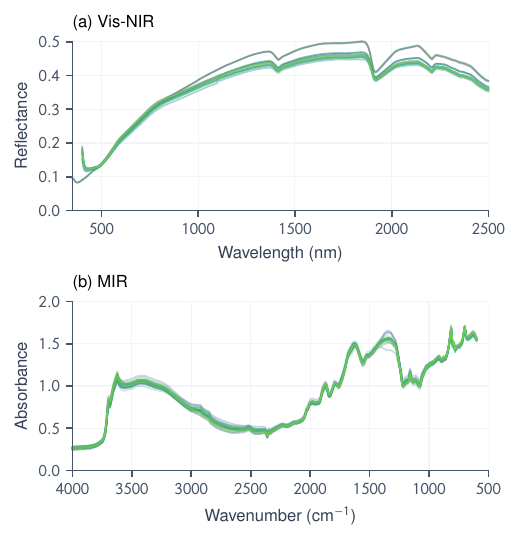}
  \caption{Mean spectra of the individual soil property prediction tasks from OSSL: (a) 16 vis-NIR tasks, and (b) 30 MIR tasks. Each line is the per-band mean spectrum across all soil samples for each target soil property. Colors differ to visually distinguish the individual tasks. Vis-NIR spectra are shown as reflectance against wavelength, and MIR spectra as absorbance against wavenumber.}
  \label{figure:datasets-spectra-ossl}
\end{figure}

For the purpose of our investigation, we employed all MIR and vis-NIR soil property prediction tasks from the OSSL for which the harmonized target soil property had at least 10{,}000 observations. This resulted in 30 MIR and 16 vis-NIR tasks, yielding a total of 46 separate soil property prediction tasks. The target soil properties span a broad range of chemical, physical, and mineralogical attributes, including, among others, organic and total carbon, total nitrogen, pH, cation exchange capacity, particle size fractions (clay, silt, sand), extractable cations and nutrients, carbonate content, bulk density, water retention parameters, and electrical conductivity. Further details are given in Supplementary Table~S2.

The MIR spectra cover the spectral range of 600--4{,}000~cm$^{-1}$ at 2~cm$^{-1}$ intervals, resulting in 1{,}701 spectral variables per sample. The vis-NIR spectra span either 350--2{,}500~nm or 400--2{,}500~nm at 2~nm intervals, depending on the spectral coverage of the underlying source libraries (e.g., the LUCAS library provides vis-NIR spectra starting from 400~nm), resulting in up to 1{,}076 spectral variables. The mean spectral signatures associated with these tasks are shown in Figure~\ref{figure:datasets-spectra-ossl}. Across the 46 tasks, dataset sizes range from 14{,}166 to 82{,}573 soil samples. The OSSL is openly accessible under permissive licenses (see \hyperref[section:code-data-availability]{Code and data availability}).

\subsection{Modeling}
\label{section:modeling}

To compare modeling strategies fairly, all soil property prediction tasks were evaluated under a common protocol: (i)~spectral preprocessing, (ii)~feature-wise standardization, (iii)~optional dimensionality reduction, and (iv)~regression, and, for skewed soil properties, (v)~a logarithmic transformation of the target. The components of this protocol are described in the following subsections.

\subsubsection{Spectral preprocessing and standardization}

All spectra were preprocessed using the standard normal variate (SNV) transformation~\citep{barnes1989standard}. SNV is a row-wise normalization that centers each spectrum to zero mean and scales it to unit standard deviation across wavelengths. SNV allows correcting for multiplicative effects of particle size and light scattering that are prevalent in diffuse reflectance spectroscopy of soil~\citep{barnes1989standard,safanelli2023interlaboratory}, and is consistently identified among the most effective preprocessing approaches for soil spectroscopy across spectral regions and modeling algorithms~\citep{safanelli2023interlaboratory,safanelli2025open,vestergaard2021evaluation}.

Following SNV, feature-wise standardization (centering to zero mean and scaling to unit variance) was applied, with standardization statistics computed on the training samples of the respective fold.

Since many soil properties are strongly right-skewed (e.g., the concentrations of carbon, nutrients, and extractable elements), we followed the modeling protocol established in the OSSL~\citep{safanelli2025open} and log-transformed the target soil properties except clay, silt, sand, and pH before model fitting. The predictions were back-transformed to the original scale before evaluation with the exponential. The same rule was applied to the field-scale datasets, where SOC and SOM were log-transformed and pH and clay were not. The transformation was applied identically to all models, including those that are designed to handle skewed targets internally, such as TabPFN.

\subsubsection{Dimensionality reduction}
\label{section:dimensionality-reduction}

Two dimensionality reduction strategies were evaluated as preprocessing steps prior to regression: principal component analysis (PCA) and partial least squares (PLS) projection.

PCA involves the decomposition of the feature matrix $\mathbf{X}$ (in our case, $I$ soil samples $\times$ $J$ spectral variables). This is commonly achieved via singular value decomposition (SVD):
\begin{equation}
  \mathbf{X} = \mathbf{R}\,\boldsymbol{\Delta}\,\mathbf{V}^\top,
  \label{eq:pca_svd}
\end{equation}
where $\mathbf{R}$ and $\mathbf{V}$ are orthonormal matrices and $\boldsymbol{\Delta}$ is a diagonal matrix of singular values. The principal components of $\mathbf{X}$ are the columns of $\mathbf{R}\boldsymbol{\Delta}$, ordered by the variance of $\mathbf{X}$ they explain. Retaining only the first $n$ components yields a reduced set of features that can be used with any regression model to predict $\mathbf{Y}$, making PCA an unsupervised dimensionality reduction method. The orthogonality of the resulting components is naturally suited to spectroscopic data, as it eliminates the multicollinearity inherent in neighboring spectral bands. However, because the decomposition is performed on $\mathbf{X}$ alone, there is no guarantee that the components explaining the most variance in $\mathbf{X}$ will be the most relevant for predicting $\mathbf{Y}$.

PLS, in contrast, searches for a set of components, commonly referred to as latent variables, that maximize the covariance between $\mathbf{X}$ and $\mathbf{Y}$. While PLS is classically computed using iterative algorithms such as Nonlinear Iterative Partial Least Squares (NIPALS), it can be expressed through SVD in a manner directly analogous to PCA~\citep{abdi2010partial}. In this formulation, the cross-covariance matrix $\mathbf{X}^\top\mathbf{Y}$ is decomposed as
\begin{equation}
  \mathbf{X}^\top\mathbf{Y} = \mathbf{W}\,\boldsymbol{\Theta}\,\mathbf{C}^\top,
  \label{eq:pls_svd}
\end{equation}
where $\mathbf{W}$ and $\mathbf{C}$ are orthonormal matrices and $\boldsymbol{\Theta}$ is a diagonal matrix of singular values reflecting the covariance between $\mathbf{X}$ and $\mathbf{Y}$. The latent variables are obtained as $\mathbf{L} = \mathbf{X}\mathbf{W}$, ordered by decreasing covariance. Retaining the first $n$ latent variables yields a reduced set of features that best capture the covariance structure between $\mathbf{X}$ and $\mathbf{Y}$. Since the decomposition involves $\mathbf{Y}$, these selected latent variables can be used directly to estimate $\hat{\mathbf{Y}}$, effectively performing PLSR. Equally, the selected latent variables can serve as features for any regression model, making PLS a supervised dimensionality reduction method that addresses potential limitations of PCA.

\subsubsection{Regression models}

Six regression models were evaluated: Linear regression, PLSR, Random Forest, Cubist, CNN, and TabPFN.

Linear regression models the relationship between dependent and independent variables using a linear equation. When multiple independent variables are used, it is referred to as multiple linear regression (MLR). MLR is a well-established baseline in pedometric modeling. A common variant of MLR employed in soil spectroscopy is principal component regression (PCR), where MLR is fitted on principal components extracted via PCA as input features~\citep[e.g.,][]{kuhn2025tracking,tavakoli2023predicting,paltseva2022prediction}.

PLSR, similarly to PCR, performs linear regression on a reduced set of components. Unlike PCR, PLSR extracts these components through simultaneous decomposition of features and the dependent variable (see Section~\ref{section:dimensionality-reduction}), effectively combining linear regression with supervised dimensionality reduction. PLSR is historically the most widely used algorithm in soil spectroscopy \citep{rossel2010using} and is still one of the strongest baselines in contemporary DSM \citep[e.g.,][]{hutengs2024enhanced,angelettidafonseca2022effect,carvalho2022combining}.

Random Forest~\citep{breiman2001random} is a randomized ensemble of decision trees regularized through bootstrap aggregation. Random Forest is the most widely adopted ML algorithm in general pedometric modeling from PSS~\citep{ding2025advancing,wadoux2020machine}. In soil spectroscopy, Random Forest is typically coupled with PCA or combined with iterative feature selection.

Cubist is a rule-based regression model that constructs a set of piecewise linear models organized as a decision tree, with optional committee boosting and instance-based nearest-neighbor correction~\citep{quinlan1993combining}. Cubist has been widely adopted for pedometric modeling in DSM \citep{minasny2008regression} with strong results in soil spectroscopy~\citep[e.g.,][]{weerasekara2026kssl,sanderman2025application,agyeman2025prediction,safanelli2025open}.

CNN is a neural network architecture in which learnable filters (commonly referred to as kernels) perform discrete convolution operations (typically implemented as spatially localized cross-correlations) by sliding across the input with a defined step size (commonly referred to as stride). CNNs are designed for data characterized by a grid-like topology, with foundational applications in image recognition~\citep{lecun2002gradient}. CNNs differ fundamentally from dimensionality reduction approaches: rather than compressing the high-dimensional spectra into fewer features, they initially increase the number of features even further, as each filter produces its own feature map. Nevertheless, they are well suited to soil spectra, as spectral bands are naturally ordered and neighboring bands are highly intercorrelated. Therefore, CNNs attracted considerable interest in soil spectroscopy, with initial applications on two-dimensional (2D) representations of spectra drawing inspiration from computer vision~\citep{padarian2019usinga}, and moving to more principled one-dimensional (1D) convolutional architectures operating directly on the spectral sequence in subsequent works~\citep{ng2019convolutional}. Accordingly, the CNN was evaluated on full spectra only and was not combined with PCA or PLS. In addition to the CNN baseline, we evaluated an ensemble of CNN models that aggregates predictions from multiple independently trained models. \citet{barkov2026modern} demonstrated that deep ensembles mitigate the inherent variability of individual artificial neural network training runs, thus allowing CNN performance to be assessed under the most favorable conditions.

TabPFN (Tabular Prior-data Fitted Network) is an in-context learning foundation model for tabular data~\citep{hollmann2025accurate,hollmann2023tabpfn}. TabPFN is pre-trained once on a large collection of synthetic tabular tasks, learning to capture the interactions among features and observations of any given tabular dataset. For a new dataset, TabPFN treats the labeled training data as a context (a ``context set'') and predicts by relating each test observation to this context set, effectively learning ``in-context'' entirely at prediction time. Because of this, TabPFN requires no dataset-specific training or hyperparameter optimization. TabPFN has demonstrated strong performance on general tabular benchmarks~\citep{grinsztajn2025tabpfn,erickson2025tabarena} and state-of-the-art results in pedometric modeling within DSM~\citep{barkov2026modern}. In the present study, we investigated TabPFN-3~\citep{grinsztajn2026tabpfn}, which, on recent general tabular benchmarks, has been shown to natively handle datasets with up to 100{,}000 training samples and nearly 2{,}000 features~\citep{erickson2025tabarena}.

\subsubsection{Experimental setup}
\label{section:experimental-setup}

The performance of the regression models and dimensionality reduction approaches was evaluated using nested cross-validation \citep{varma2006bias}. For both the LimeSoDa and OSSL datasets, 5-fold outer cross-validation was employed, with folds created by random splitting. The same fold partitions were used across all model and dimensionality reduction configurations within each dataset. Hyperparameters of each model were optimized within an inner validation loop. When dimensionality reduction was applied, the number of retained components for PCA or latent variables for PLS was treated as an additional hyperparameter~\citep{vanleeuwen2026effect}. For the field-scale LimeSoDa datasets, inner validation used 5-fold cross-validation. For the large-scale OSSL datasets, an inner holdout split (20\% of the outer training set) was used instead to reduce computational cost. Hyperparameters were optimized using Bayesian optimization with the Tree-structured Parzen Estimator (TPE) algorithm~\citep{bergstra2011algorithms} as implemented in Optuna~\citep{akiba2019optuna}. Further details on the hyperparameter search spaces and training configurations are provided in Supplementary Table~S7. All experiments were run on dedicated hardware, with linear regression, PLSR, and Cubist run on a single core of an AMD EPYC 7742 central processing unit (CPU), Random Forest on two AMD EPYC 7742 CPUs with trees parallelized across 128 cores, and CNN and TabPFN on an NVIDIA A100 80\,GB graphics processing unit (GPU). To support reproducibility, source code is provided openly, together with all software dependencies and their exact versions used in the experiments (see \hyperref[section:code-data-availability]{Code and data availability}).

Predictive performance on each task was assessed using the root mean squared error (RMSE),
\begin{equation}
  \mathrm{RMSE} = \sqrt{\frac{1}{n}\sum_{i=1}^{n}(y_i - \hat{y}_i)^2},
  \label{eq:rmse}
\end{equation}
where $y_i$ and $\hat{y}_i$ denote the observed and predicted values for the $i$-th test sample, and $n$ is the number of samples in the held-out test fold. RMSE was used both as the optimization objective during hyperparameter optimization and as the metric for assessing generalization on the outer test folds.

Since the evaluated tasks involved soil properties with different scales and units, ranking based on RMSE was employed~\citep{rosset2005rankingbased,nariya2023paired}. Mean and median ranks of different models and the dimensionality reduction configurations served as the primary measure of relative performance.

To investigate the magnitude of differences between different approaches, we additionally investigate the relative RMSE differences of models in relation to the best achieved RMSE of each task.

For further analysis, Supplementary Figures~S1--S4 include the ranking based on the mean absolute error (MAE), and Supplementary Tables~S3--S6 include the coefficient of determination (R$^2$), the ratio of performance to interquartile range (RPIQ), and Lin's concordance correlation coefficient (CCC) aggregated across the investigated groups.

\subsubsection{Spectral regions and spectral mechanisms}
\label{section:grouping}

Soil properties differ in how they relate to the spectra~\citep{stenberg2010visible}. Some properties correspond to soil constituents with absorption features of their own, such as organic matter, clay minerals, carbonates, and iron oxides, whereas others, such as pH, cation exchange capacity, and nutrients, have no direct spectral response and can only be predicted through their covariation with spectrally active constituents~\citep{stenberg2010visible}. To examine whether the relative performance of the modeling strategies depends on this spectral mechanism, we investigate the results in both spectral regions across two groups of target properties: (1) ``direct'', for spectrally active constituents with established absorption features, and (2) ``indirect'', for properties without absorption features of their own that are predicted through their covariation with spectrally active constituents. In the field-scale datasets, SOC/SOM and clay belong to the direct group and pH to the indirect group. Detailed information on the soil property groups of the large-scale spectral library tasks is included in Supplementary Table~S2.

\section{Results}

\subsection{Overall ranking of models and dimensionality reduction strategies}
\label{section:overall-ranking}

\begin{figure*}[pos=htbp]
  \centering
  \includegraphics[width=\textwidth]{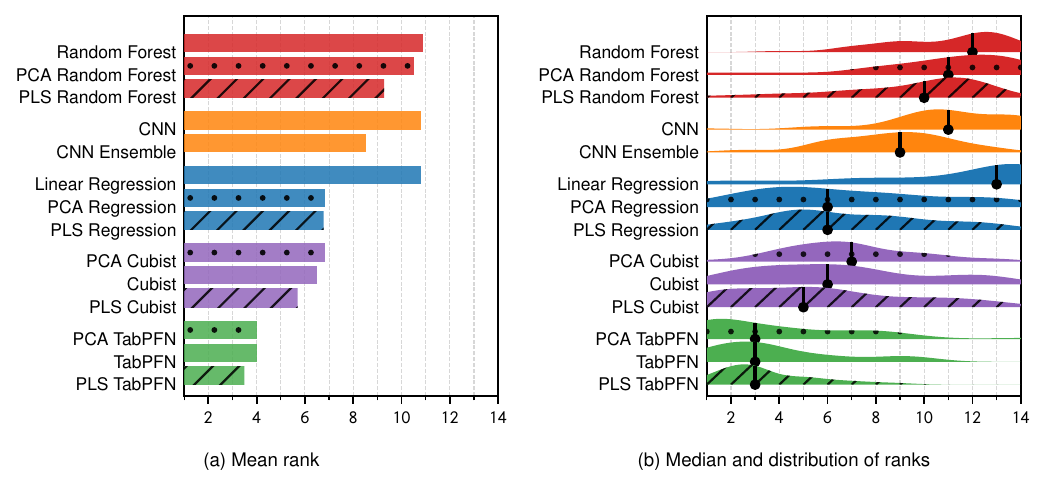}
  \caption{Ranking of 14 model and dimensionality reduction configurations evaluated across 39 field-scale soil property prediction tasks from the LimeSoDa dataset collection, showing: (a) mean ranks, and (b) distribution of ranks across tasks, with the large dot marking the median rank. Ranking is based on RMSE. Lower ranks indicate better predictive performance.}
  \label{fig:results-limesoda}
\end{figure*}

\begin{figure*}[pos=htbp]
  \centering
  \includegraphics[width=\textwidth]{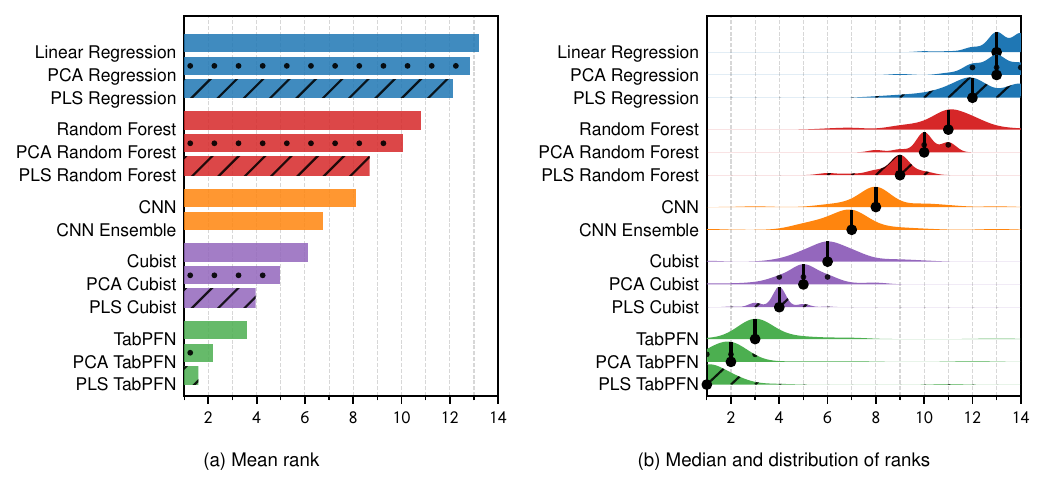}
  \caption{Ranking of 14 model and dimensionality reduction configurations evaluated across 46 soil property prediction tasks from the Open Soil Spectral Library (OSSL), showing: (a) mean ranks, and (b) distribution of ranks across tasks, with the large dot marking the median rank. Ranking is based on RMSE. Lower ranks indicate better predictive performance.}
  \label{fig:results-ossl}
\end{figure*}

Results are reported separately for the 39 field-scale LimeSoDa tasks and the 46 OSSL tasks. Figures~\ref{fig:results-limesoda} and~\ref{fig:results-ossl} summarize the mean and median ranks and rank distributions of the 14 model and dimensionality reduction combinations. Across both dataset groups, a similar ordering of models and their variants emerged.

TabPFN delivered the strongest performance overall. On both LimeSoDa and OSSL, the three TabPFN variants outperformed all other configurations, with PLS TabPFN ranked first in both settings. On the field-scale datasets, the three TabPFN variants were clearly ahead of the remaining configurations and close to each other. On the large-scale spectral library, PLS TabPFN showed the most evident advantage, and was the best configuration on 31 of the 46 tasks. Notably, even without a separate dimensionality reduction step, TabPFN achieved the best or near-best results in both dataset groups.

Cubist was the strongest classical baseline. PLS Cubist ranked directly behind the TabPFN variants in both dataset groups and clearly improved over Cubist on full spectra and PCA Cubist. Although PLS Cubist did not outperform PLS TabPFN, it achieved a lower RMSE than TabPFN on full spectra on 9 of the 46 tasks.

CNN showed moderate performance overall, with better results on the large-scale spectral library. The CNN ensemble variant consistently outperformed a single CNN and showed a reduced spread of ranks overall, especially noticeable on the field-scale datasets. Random Forest did not perform competitively in either dataset group. PLS latent variables as inputs improved Random Forest relative to full spectra and PCA components, but not sufficiently to close the gap to the best-performing models.

The linear models were competitive at the field scale, where PLSR and PCR ranked close to Cubist on full spectra, whereas on the large spectral library all linear models ranked last.

A consistent ordering was also observed among dimensionality reduction strategies. PLS was the most effective strategy across both dataset groups. PLSR was the best-performing linear model and outperformed both linear regression and PCR. The benefit of PLS as a feature extraction method extended to the other models as well. Using PLS latent variables instead of full spectra or PCA components improved Cubist, Random Forest, and TabPFN in most configurations. For instance, PLS Cubist achieved a lower RMSE than Cubist on full spectra on 26 of the 39 field-scale and 43 of the 46 large-scale spectral library tasks, and PLS TabPFN a lower RMSE than TabPFN on full spectra on 20 of the 39 field-scale and 44 of the 46 large-scale spectral library tasks. PCA provided some improvement over full spectra, especially on the large-scale spectral library, but, overall, these improvements were smaller and less consistent.

\subsection{Relative performance across tasks}
\label{section:relative-performance}

\begin{figure*}[pos=htbp]
  \centering
  \includegraphics[width=\textwidth]{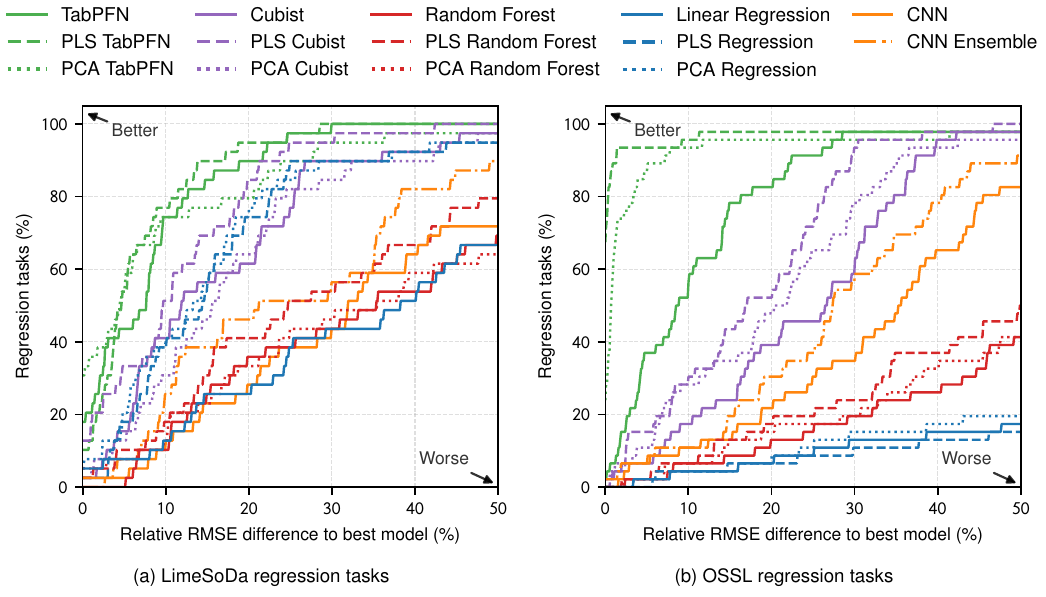}
  \caption{Relative performance of the 14 model and dimensionality reduction configurations on (a) the 39 field-scale soil property prediction tasks from LimeSoDa and (b) the 46 large-scale soil property prediction tasks from OSSL. For each task, the RMSE of every configuration is expressed as the relative difference to the RMSE of the best-performing configuration on that task (x-axis). Lines show the percentage of tasks (y-axis) on which a configuration falls within a given relative RMSE difference of the best model. Lines toward the upper left indicate configurations that perform at or near the best achievable RMSE on most tasks.}
  \label{fig:results-performance-profile}
\end{figure*}

Relative performance on field-scale data (Figure~\ref{fig:results-performance-profile}a) shows that the TabPFN variants stayed consistently close to the best achievable RMSE. PLS TabPFN showed the smallest mean relative gap to the best achievable RMSE (6.7\%), stayed within 10\% of the best achievable RMSE on 77\% of the tasks, and within 20\% on 37 of the 39 tasks. TabPFN on full spectra and PCA TabPFN stayed within 10\% of the best achievable RMSE on 74\% of the tasks each, with PCA TabPFN showing a wider spread (mean gap of 8.9\%). PLS Cubist remained within 10\% of the best model on about half (51\%) of the tasks. In paired terms, PLS TabPFN achieved a lower RMSE than PLS Cubist on 30 of the 39 tasks, with a median RMSE reduction of 4.3\%, a lower RMSE than PLSR on 32 tasks (median reduction 6.2\%), and a lower RMSE than the CNN ensemble on 35 tasks (median reduction 11\%).

On the large-scale spectral library (Figure~\ref{fig:results-performance-profile}b), the separation between the best configurations and the remaining models was far more pronounced. PLS TabPFN was the best configuration on 67\% of the tasks (31 of 46), and showed RMSE within 1\% of the best on 89\% and within 10\% of the best on 96\% of the tasks. PCA TabPFN followed closely, with an RMSE within 10\% of the best on 93\% of the tasks. TabPFN on full spectra showed a larger gap, but still achieved an RMSE within 10\% of the best on 54\% of the tasks, more often than any configuration outside the investigated TabPFN variants. PLS Cubist remained the strongest classical configuration, with an RMSE within 20\% of the best achievable one on about half of the tasks (52\%). In paired terms, PLS TabPFN reduced the RMSE relative to PLS Cubist on 44 of the 46 tasks, by a median of 14\%, relative to the CNN ensemble on 45 of the 46 tasks by a median of 21\%, and relative to PLSR on all 46 tasks by a median of 49\%.

\subsection{Results by spectral region and spectral mechanism of the soil property}
\label{section:grouped-results}

\begin{figure*}[pos=htbp]
  \centering
  \includegraphics[width=0.78\textwidth]{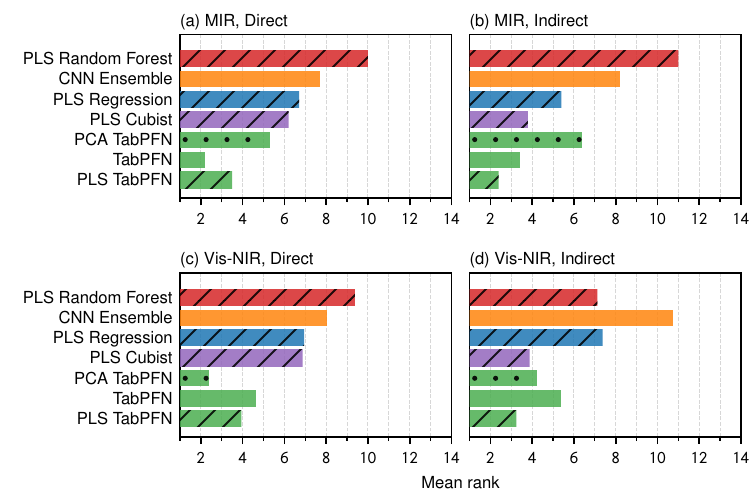}
  \caption{Mean rank among all 14 configurations on the field-scale LimeSoDa tasks. The figure shows all three TabPFN variants and the best-performing variant of each other model. Results are grouped by spectral region and spectral mechanism group of the target property: (a) MIR, direct (SOC/SOM and clay); (b) MIR, indirect (pH); (c) vis-NIR, direct (SOC/SOM and clay); and (d) vis-NIR, indirect (pH). Ranking is based on RMSE. Lower ranks indicate better predictive performance.}
  \label{fig:grouped-limesoda}
\end{figure*}

\begin{figure*}[pos=htbp]
  \centering
  \includegraphics[width=0.78\textwidth]{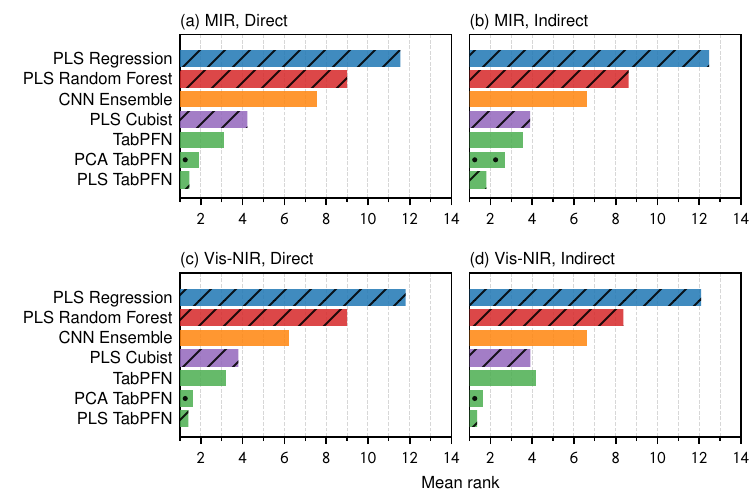}
  \caption{Mean rank among all 14 configurations on the OSSL large-scale spectral library tasks. The figure shows all three TabPFN variants and the best-performing variant of each other model. Results are grouped by spectral region and spectral mechanism group of the target property: (a) MIR, direct; (b) MIR, indirect; (c) vis-NIR, direct; and (d) vis-NIR, indirect. Ranking is based on RMSE. Lower ranks indicate better predictive performance.}
  \label{fig:grouped-ossl}
\end{figure*}

When investigating the results according to the spectral region and the spectral mechanism group of the target property on the field-scale datasets (Figure~\ref{fig:grouped-limesoda}), TabPFN variants ranked first in every group, although the best variant differed slightly between spectral regions. In many cases, Cubist was also competitive. For the indirect property (pH), PLS TabPFN ranked first in both the MIR and the vis-NIR region (Figure~\ref{fig:grouped-limesoda}b,d). When predicting pH from vis-NIR spectra, PLS TabPFN (mean rank of 3.3) was closely followed by PLS Cubist (mean rank of 3.9), while TabPFN on full spectra showed worse results than both (mean rank of 5.4).

When considering the spectrally active properties (SOC/SOM and clay) at the field scale, TabPFN applied directly to the full spectra ranked first in the MIR region (Figure~\ref{fig:grouped-limesoda}a; mean rank of 2.2, compared to 3.5 for PLS TabPFN). On the vis-NIR datasets, PCA TabPFN ranked first for SOC/SOM and clay (Figure~\ref{fig:grouped-limesoda}c; mean rank of 2.4, compared to 3.9 for PLS TabPFN). At the field scale, PLS latent variables were thus most useful for the indirectly predicted pH, whereas for the spectrally active properties the full spectra (MIR) or principal components (vis-NIR) were better.

On the large spectral library (Figure~\ref{fig:grouped-ossl} and Supplementary Tables~S5 and~S6), PLS TabPFN ranked first in all four groups, with mean ranks between 1.4 and 1.8, and PCA TabPFN ranked second in all four groups. TabPFN on full spectra ranked third for the direct properties in both spectral regions and for the indirect properties from MIR spectra (mean ranks of 3.1--3.6), ahead of PLS Cubist (3.8--4.2), whereas for the indirect properties from vis-NIR spectra, PLS Cubist (3.9) moved slightly ahead of TabPFN on full spectra (4.2; Figure~\ref{fig:grouped-ossl}d). The advantage of PLS TabPFN over PLS Cubist was larger in the vis-NIR region, where PLS TabPFN reduced the RMSE relative to PLS Cubist by a median of 22\% for the direct and 17\% for the indirect properties, than in the MIR region, where the corresponding median reductions were 13\% and 10\%. For the indirect properties, PLS TabPFN ranked first in both spectral regions, followed by PCA TabPFN, and achieved a lower RMSE than PLS Cubist on 30 of the 32 indirect tasks. The advantage of TabPFN therefore also held for properties that are predicted only through their covariation with spectrally active constituents.

\subsection{Computational cost}
\label{section:computational-cost}

\begin{figure*}[pos=htbp]
  \centering
  \includegraphics[width=\textwidth]{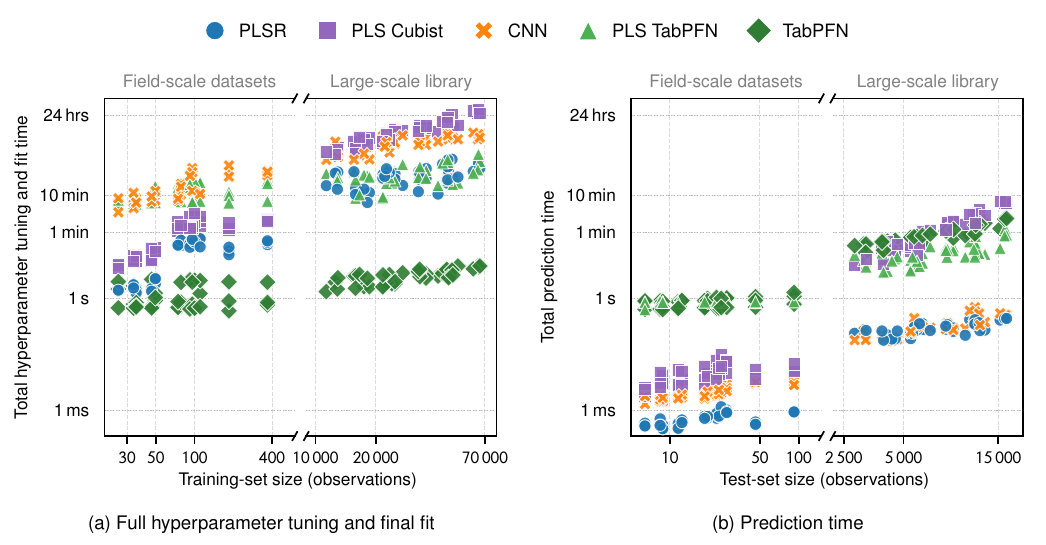}
  \caption{Runtime scaling of the benchmarked models. Left: full hyperparameter tuning plus final fit, as a function of training-set size. Right: prediction time as a function of test-set size. Axes are log--log. Each point corresponds to one task, reported as the median wall-clock time across its cross-validation folds. CNN and TabPFN were run on a GPU, while the remaining models were run on a CPU.}
  \label{fig:runtime}
\end{figure*}

The wall-clock training and prediction times for each evaluated task are shown in Figure~\ref{fig:runtime}. The figure includes a selection of the best-performing models in this study together with commonly used ones: PLSR, CNN, TabPFN, PLS TabPFN, and PLS Cubist.

The left panel in Figure~\ref{fig:runtime} shows the median wall-clock time for model training, including hyperparameter optimization. Training time increased with the size of the training set for all models and configurations except TabPFN applied directly to the spectra. Since TabPFN requires no hyperparameter optimization, its training included only a single model fit (which, in TabPFN specifically, corresponds to just processing and caching the context set). As a result, TabPFN had the lowest training time of all configurations, and it remained almost constant with respect to the training set size. Among the remaining models, PLS Cubist and CNN had the highest training times, reaching up to approximately 34\,h and 20\,h of median wall-clock time per fold, respectively, on the largest tasks.

The right panel in Figure~\ref{fig:runtime} shows the median wall-clock time for predicting the test set, after the model with the best hyperparameters had been selected and fitted. CNN, despite having one of the highest training times, predicted in well under a second even for the largest test sets, taking up to 6\,ms on field-scale and up to 590\,ms on large-scale datasets. The same held for PLSR, with up to 1.2\,ms on field-scale and up to 295\,ms on large-scale datasets. TabPFN, by contrast, showed noticeably higher prediction times that scaled with dataset size. On field-scale tasks its prediction time still remained below 1.5\,s, but on large-scale tasks it grew to up to 2.4\,min of median wall-clock time per fold. Reducing the spectra to PLS latent variables lowered TabPFN prediction time by about two thirds (up to 1.3\,min on the largest tasks), while the scaling with dataset size persisted. Interestingly, PLS Cubist showed prediction times of a similar order of magnitude, but with a steeper scaling. On the large-scale library, the prediction times of TabPFN, PLS TabPFN, and PLS Cubist all grew approximately quadratically with test-set size.

\section{Discussion}

Vis-NIR and MIR spectroscopy holds promise to make large-scale physicochemical analysis faster, cheaper, and less laborious than traditional laboratory procedures~\citep{nocita2015soilspectroscopy}, provided that the predictive models behind it are both accurate and methodologically sound~\citep[cf.][]{mcbride2022estimating}. Our findings contribute to this. (1)~The tabular foundation model, TabPFN, showed the best predictive performance overall, even on large-scale spectral library soil property prediction tasks (see Section~\ref{section:discussion-scaling}). While better accuracy is a worthy aim, and remains a precondition for the wider adoption of soil spectroscopy~\citep{mcbride2022estimating}, prediction is not the only purpose of modeling in soil science~\citep{wadoux2020machine,minasny2024soil}, and our findings go further. (2)~TabPFN required no dataset-specific training or hyperparameter optimization, removing some of the sources of methodological bias that commonly weigh on soil spectroscopy practitioners (see Section~\ref{section:discussion-scaling}). (3)~Many soil properties have no direct spectral response, and a model thus has to infer them from their covariation with constituents that are spectrally active. In our findings, the advantage of the tabular foundation model held for properties with and without a direct spectral response and across both vis-NIR and MIR spectra (see Section~\ref{section:discussion-groups}). (4)~PLS proved an effective supervised feature extractor for any regression model, not only for PLSR, and showed a good synergy with TabPFN (see Section~\ref{section:discussion-pls}). (5)~The inference cost of TabPFN was higher, but is not a critically limiting barrier for soil spectroscopy (see Section~\ref{section:discussion-cost}). Finally, (6)~our conclusions rest on 85 soil property prediction tasks from open benchmark datasets that differ in operational scale, spectrometer instruments, and soil properties, and are thus subject to fewer biases than typical single-dataset comparisons~\citep{schmidinger2025limesoda}.

\subsection{Do tabular foundation models scale to large soil spectral libraries?}
\label{section:discussion-scaling}

In previous exploratory works, \citet{barkov2026modern} and \citet{huang2026zeroshot} showed that TabPFN delivers strong results for soil spectroscopy, with \citet{barkov2026modern} proposing it as a new default for field-scale datasets and \citet{huang2026zeroshot} taking an initial step toward exploring how it scales further. Nonetheless, the question of whether it would scale to large spectral libraries still stands. This is a fair question, especially given the scale at which these models were first developed: prior-data fitted networks were introduced on small synthetic tabular tasks of 100 observations \citep{muller2022transformers}, and the first version of TabPFN was trained and evaluated on datasets of only up to roughly 1{,}000 observations \citep{hollmann2023tabpfn}, bringing the common expectation that tabular foundation models are mostly constrained to settings with limited data availability. Established deep learning approaches such as CNNs, by contrast, have shown great potential in soil spectroscopy, with their main limitation being poor performance on small sample sizes, which further calls into question whether changing the modeling approach to a tabular foundation model is necessary at all when large amounts of data are available. In our study, however, TabPFN excelled across fundamentally different scales, including soil property prediction tasks involving tens of thousands of ground-truth soil samples, extending recent DSM evidence beyond the field-scale applications of \citet{barkov2026modern} and the national soil spectral library examined by \citet{huang2026zeroshot}. Such scalability did not happen by chance: scaling in-context learning beyond small datasets was identified early on as both a practical need and a technical challenge, and became an important line of research~\citep{qu2025tabicl}. Recent TabPFN versions, such as TabPFN-3 investigated in this work, followed this path by scaling their synthetic pre-training to substantially larger sample and feature counts and by implementing architectural decisions such as feature subsampling across underlying ensemble members, row chunking, and a reduced key--value cache~\citep{grinsztajn2025tabpfn,grinsztajn2026tabpfn}. Tabular foundation models thus now appear to be a viable option for spectroscopic modeling across all operational scales.

Moreover, the advances in recent tabular foundation models move us toward modeling directly on the spectra, without the dimensionality reduction step previously proposed as a workaround for the feature-size limits of earlier versions, both in soil spectroscopy \citep{barkov2026modern} and in general tabular ML \citep{ye2025closer}. Practitioners can thus employ TabPFN on spectroscopic data with minimal preprocessing while still achieving state-of-the-art predictive performance. This also reframes the role of dimensionality reduction: rather than a workaround imposed by model constraints, strong dimensionality reduction can instead be employed deliberately and contextually to obtain better results, as we discuss next.

TabPFN needed neither dataset-specific training nor hyperparameter optimization. This is convenient, but, more importantly, it leaves soil spectroscopy practitioners less room for error. Modeling in soil spectroscopy can combine many different, and frequently interdependent, choices: spectral preprocessing, feature selection (frequently iterative) and dimensionality reduction, model-specific hyperparameter optimization, post-processing steps, and more. Each of these choices, unless considered carefully, is a potential source of data leakage or overfitting~\citep{correa2025overfitting,mcbride2022estimating}, and thereby of overly optimistic conclusions. Fewer decisions make the modeling easier to apply correctly, and allow the focus to shift from tuning models to the soil science questions the predictions are meant to answer~\citep{minasny2024soil}.

\subsection{Did TabPFN perform equally well across spectral regions and spectral mechanisms?}
\label{section:discussion-groups}

Many soil properties are not spectrally active, and the predictability of such indirect properties is bounded by the strength of their covariation with spectrally active constituents such as organic matter, clay minerals, and carbonates~\citep{stenberg2010visible,irving2026infrared}. Likewise, the two spectral regions differ in how specific the available information is.

The grouped results show that the advantage of TabPFN did not depend on the spectral region or on the spectral mechanism of the target property. A TabPFN variant ranked first in every combination of spectral region and property group at both operational scales, for the spectrally active constituents as well as for the properties that are predicted indirectly through their covariation with these constituents. PLS TabPFN was the best configuration in all groups of the large-scale spectral library and for the indirectly predicted pH at the field scale, whereas for the spectrally active properties of the field-scale datasets, TabPFN on full spectra (MIR) and PCA TabPFN (vis-NIR) performed better. TabPFN thus appears viable for both spectral regions and both groups of soil properties. The margins, however, differed between the regions: on the spectral library, the gains of PLS TabPFN over the strongest classical baseline were larger for vis-NIR than for MIR spectra. In the MIR region the strong models were often close to each other, whereas in the vis-NIR region the choice of the model mattered more.

Nevertheless, while TabPFN seems to be a good starting point for soil spectroscopy practitioners, it is important to consider which model, and which surrounding pipeline, best fits a given task. Sometimes, simpler models suffice. Some spectrally active soil properties were predicted almost equally well by all strong models: for organic and total carbon from MIR spectra, the RMSEs of the TabPFN variants (1.00--1.06\% w/w) and of the Cubist variants (1.15--1.22\% w/w) were all within 0.2\% w/w of each other. For other soil properties, the differences were more critical: for carbonates from vis-NIR spectra, PLS TabPFN reached an RMSE of 2.71\% w/w, compared to 3.98\% w/w for PLS Cubist (47\% higher) and 4.23\% w/w for the CNN ensemble (56\% higher), and for clay from vis-NIR spectra an RMSE of 4.69\% w/w, compared to 6.09\% w/w for PLS Cubist (30\% higher). Conversely, some indirectly predicted properties, such as Mehlich-3 phosphorus and KCl-extractable manganese, were poorly predicted by every model, as the spectra carry little information about these properties in the first place. The hierarchy of the models thus held across spectral regions and mechanisms, but the magnitude of the improvements, and whether they matter for the practitioner's application at hand, still have to be judged case by case.

\subsection{On the role of PLS in soil spectroscopy}
\label{section:discussion-pls}

Although TabPFN applied directly to spectra already outperformed established modeling approaches, our results show that further consistent improvements remain attainable, most immediately through dimensionality reduction, with the strongest and most consistent gains coming from supervised, target-aware PLS. This is an intuitive insight: PLSR has persisted in soil spectroscopy for decades, both as a predictive model \citep{soriano-disla2014performance,viscarrarossel2006visible,gyawali2025measuring,reyes2024spectral,pace2024soil} and as an interpretability tool whose weights can reveal the mineral and organic soil constituents that correlate with the target property~\citep{janik1995characterization}. Our findings suggest that the strength behind this longevity can be carried forward even as the field adopts tabular foundation models.

PLSR was the strongest linear model across both dataset groups, consistent with its long track record, but more importantly, the advantages of its PLS decomposition extended well beyond PLSR itself. Across Cubist, Random Forest, and TabPFN, pairing the regression model with extracted PLS latent variables consistently improved predictions relative to both full spectra and PCA components. Combining this supervised feature extraction with TabPFN produced the overall best results in our study. Ultimately, PLS proved to be an effective step for maximizing the capabilities of modern regression algorithms in soil spectroscopy.

\subsection{Does TabPFN shift the computational cost to the prediction step?}
\label{section:discussion-cost}

Besides showing the best predictive performance, TabPFN effectively eliminated the cost of model training. As expected, this comes with trade-offs at the prediction stage. First, as an artificial neural network, TabPFN generally requires a GPU for optimal performance. GPUs, being massively parallel processors, offer advantages such as efficient batch processing, but also present operational constraints in settings where such hardware is not readily available. Moreover, GPU computation is energy-intensive and can carry a substantial carbon footprint. Further, although TabPFN removes the need for dataset-specific training, it still requires a labeled context set of ground-truth observations to make predictions. This shifts the computational cost of modeling to the prediction stage, where it grows with the size of the dataset. Most importantly, the attention mechanism in TabPFN incurs a computational complexity of $O(n^2 + m^2)$ with respect to sample size $n$ and feature count $m$~\citep{hollmann2025accurate}. Both terms are unfavorable when working with large spectroscopy libraries, as was directly reflected in the wall-clock timings of our experiments (Figure~\ref{fig:runtime}).

However, placed in the context of models commonly used in soil spectroscopy, these costs appear neither critical nor unusual. The reliance on a GPU is already common among deep learning approaches such as CNNs, and the scaling inference cost is likewise not unique to TabPFN. Cubist, a long-established model in soil spectroscopy, similarly showed near-quadratic scaling of prediction time with test-set size in our experiments. This scaling is, of course, intuitive, and attributable to Cubist's optional instance-based nearest-neighbor correction: for each test observation, Cubist locates the nearest training instances and adjusts the rule-based prediction by the model's residuals at those neighbors~\citep{quinlan1993combining}. While this mechanism in Cubist aligns well with spectroscopy, where each observation carries a characteristic spectral signature and prediction by similarity to reference spectra is itself a long-standing and effective strategy~\citep{ramirez2013distance}, it ties prediction cost to the size of the training set, much like TabPFN's reliance on a context set. In our experiments, PLS Cubist and PLS TabPFN showed prediction times of comparable magnitude and similarly steep scaling: the high inference cost of high-performing models in spectroscopy is not a new barrier introduced by tabular foundation models.

Whether these costs are justified depends strongly on the needs of the soil spectroscopy practitioner. Conventional modeling incurs its cost during model development, which can be limiting enough to be deliberately restricted~\citep[e.g.,][]{ramirez2026when} and has to be repeated for different data and different soil properties~\citep[e.g.,][]{huang2026zeroshot}. TabPFN prediction time reached minutes on the largest soil property prediction tasks, which is neither a new nor an especially restrictive barrier for a typical application. The hardware requirement is a more restrictive limitation, as a GPU is ideally required. Nevertheless, both appear negligible against the soil sample acquisition, preparation, sensing, and, above all, the laboratory analyses that spectroscopy is meant to replace~\citep{nocita2015soilspectroscopy} or assist with~\citep{schmidinger2026rejector}. Moreover, TabPFN is, of course, complex in its internals, but it is simple in its use. This ease of use, which requires fewer decisions (see Section~\ref{section:discussion-scaling}), weighs further in favor of tabular foundation models, even at higher costs. Finally, the inference cost of tabular foundation models is set only to decrease: pairing TabPFN with PLS already lowered its prediction time by about two thirds in our experiments. More efficient architectures are being developed~\citep[e.g.,][]{qu2026tabiclv2}, and further improvements can be made by retrieving only relevant local context, akin to memory-based learning~\citep{ramirez2026rethinking}.

\subsection{Practical guidance, outlook, and future directions}

Taken together, our results indicate that modern tabular foundation models such as TabPFN are a strong default choice for soil spectroscopy. Several practical considerations should nonetheless be taken into account. The most important is, as discussed above, the computational cost at inference time. Practitioners should remain mindful of it when working with very large datasets. The choice of a dimensionality reduction strategy also carries practical implications. PLS is target-specific, so a specific projection must be fitted for each new soil property, making PCA a more flexible choice at times. Moreover, other feature extraction techniques are also applicable in soil spectroscopy, and PCA and PLS themselves can be made more robust to outliers~\citep{croux2000principal,serneels2005partial}. PLS TabPFN is thus a good default choice, but the dimensionality reduction strategy should still be considered on a case-by-case basis.

Another important consideration is that, in our experimental setup, training and test samples were drawn from the same underlying population, corresponding to an independent and identically distributed (IID) evaluation. Ensuring that a spectral library is representative of its intended prediction targets is, of course, a guiding requirement when building such libraries~\citep{dorantes2022calibration}. In practice, however, many chemometric applications eventually face out-of-distribution (OOD) conditions: spectral libraries can fail to capture site-specific variability~\citep{dorantes2022calibration}, and models trained on a spectral library are sometimes applied to unseen sites and regions~\citep{ng2022spike} or to spectra acquired with different instruments and in different laboratories~\citep{safanelli2023interlaboratory}. Because such distribution shifts are well known to degrade predictive performance~\citep[e.g.,][]{sun2026selfsupervised,hateffard2024evaluating}, whether the advantages of tabular foundation models observed here persist under OOD conditions requires further evaluation. In our case specifically, such an evaluation could account for the spatial autocorrelation between neighboring samples and for samples taken at different depths of the same soil profile, such as the horizons in the OSSL. Early evidence from general tabular ML suggests that tabular foundation models remain a leading model family in the OOD setting~\citep{purucker2026beyond}, and we regard a systematic OOD evaluation of tabular foundation models for soil spectroscopy as the most immediate extension of the present work.

Our findings also leave many concrete avenues for future work. Most tabular foundation models provide a full predictive distribution at prediction time, so uncertainty estimation is possible at no additional computational cost \citep{huang2026zeroshot,schmidinger2026kriging}, making it possible to combine spectroscopic and laboratory measurements according to the strengths of each approach \citep{schmidinger2026rejector}. Another opportunity lies in fine-tuning on real soil spectroscopy data \citep{garg2025realtabpfn}. Finally, the landscape of tabular foundation models evolves rapidly \citep{erickson2025tabarena}. During the preparation of this manuscript, further improvements to TabPFN were already introduced, including TabPFN-3.5~\citep{jaeger2026tabpfn35}. Other models, such as TabICLv2~\citep{qu2026tabiclv2} and NVIDIA Kumo Tabular, also show strong performance and may become attractive options for spectroscopic modeling owing to their open licensing compared to TabPFN. Overall, the benefits of tabular foundation models for soil spectroscopy are already evident, and with many avenues still left to explore, the field can be expected to advance considerably in the near future.

\section{Conclusions}

Tabular foundation models emerge as the best-performing model family for soil property prediction from vis-NIR and MIR spectroscopy. Across 85 soil property prediction tasks from openly accessible field-scale and large-scale benchmark datasets, TabPFN consistently outperformed established approaches for spectroscopic pedometric modeling, including PLSR, Cubist, Random Forest, and CNN. Importantly, this advantage extended to the large soil spectral library setting, where tasks involved tens of thousands of soil samples, thereby extending recent evidence for tabular foundation models in pedometrics beyond field-scale DSM. The strong performance of TabPFN held in both spectral regions and for soil properties with a direct spectral response as well as for those predicted indirectly through their covariation with spectrally active constituents, with the largest gains over the classical baselines for vis-NIR spectra on the large-scale spectral library. The best results were obtained with PLS TabPFN. At the same time, TabPFN applied directly to full spectra already outperformed all classical baselines, showing that explicit dimensionality reduction is not strictly required for strong predictive performance in soil spectroscopy and offering practitioners a high-performing workflow with minimal preprocessing.

When opting for dimensionality reduction, the choice of strategy proved to matter. Target-aware dimensionality reduction achieved with PLS was most effective and consistently outperformed PCA across both field-scale DSM and large soil spectral libraries. This benefit was not limited to linear models, but extended to Cubist, Random Forest, and TabPFN.

Among the conventional approaches, Cubist was the strongest overall baseline when paired with PLS, while PLSR remained the strongest linear model. CNN showed moderate performance, benefited from larger datasets and ensembling, but ultimately did not surpass tabular foundation models.

Taken together, our results provide straightforward guidance for pedometric practice. We recommend tabular foundation models as the default modeling strategy for soil spectroscopy across operational scales. When predictive accuracy is the main objective, pairing TabPFN with PLS latent variables could bring further improvements. When a straightforward workflow is preferred, TabPFN on full spectra provides strong performance without dataset-specific training or model-specific hyperparameter optimization. Overall, our results further position tabular foundation models as the strong default choice for soil spectroscopy.

\section*{Code and data availability}
\phantomsection
\label{section:code-data-availability}

The source code supporting the experiments is open and publicly available at \url{https://github.com/slavabarkov/soilscope}. All the data used in this study are publicly available in the LimeSoDa repository \citep{schmidinger2025limesoda} at \url{https://doi.org/10.5281/zenodo.14932572} and in the OSSL repository \citep{safanelli2025open} at \url{https://doi.org/10.5281/zenodo.5759693}.

\section*{Acknowledgements}
\phantomsection
\label{section:acknowledgements}

This work was supported by the Lower Saxony Ministry of Science and Culture (MWK), via the zukunft.niedersachsen program of the Volkswagen Foundation (ZN4072). Compute resources were supported by the Deutsche Forschungsgemeinschaft (DFG, German Research Foundation) project number 456666331.

\section*{Supporting information}
\phantomsection
\label{section:supporting-information}

A supporting information file for this article can be found at \href{https://doi.org/10.1016/j.geoderma.2026.118138}{https://doi.org/10.1016/j.geoderma.2026.118138}. The supporting information file contains the supplementary tables and figures listed as follows. \textbf{Table~S1}: Detailed information on the 13 field-scale soil spectroscopy datasets. \textbf{Table~S2}: Detailed information on the 46 large-scale soil property prediction tasks. \textbf{Figure~S1}: MAE-based ranking of the models and dimensionality reduction approaches across the field-scale tasks. \textbf{Figure~S2}: MAE-based ranking of the models and dimensionality reduction approaches across the spectral library tasks. \textbf{Figure~S3}: MAE-based ranking per spectral region and spectral mechanism group across the field-scale tasks. \textbf{Figure~S4}: MAE-based ranking per spectral region and spectral mechanism group across the spectral library tasks. \textbf{Table~S3}: Extended results on the field-scale MIR tasks. \textbf{Table~S4}: Extended results on the field-scale vis-NIR tasks. \textbf{Table~S5}: Extended results on the spectral library MIR tasks. \textbf{Table~S6}: Extended results on the spectral library vis-NIR tasks. \textbf{Table~S7}: Hyperparameter search spaces. \textbf{Figure~S5}: Number of principal components and PLS latent variables selected on the spectral library tasks.

\bibliographystyle{cas-model2-names}
\bibliography{references}

\end{document}